\documentclass[runningheads]{llncs}
\usepackage[hidelinks]{hyperref} 
\usepackage[T1]{fontenc}
\usepackage{graphicx}
\usepackage{multirow}%
\usepackage{mathrsfs}%
\usepackage{textcomp}%
\usepackage{booktabs}%
\usepackage{algorithm}%
\usepackage{algorithmicx}%
\usepackage{algpseudocode}%
\usepackage{listings}%
\usepackage{enumitem}
\usepackage{hyperref}
\usepackage{subcaption}
\usepackage{orcidlink}
\usepackage{cite}
\usepackage{enumitem}
\setlist{nosep}

\newcommand{\repthanks}[1]{\textsuperscript{\ref{#1}}}
\makeatletter
\patchcmd{\maketitle}
{\def\thanks}
{\let\repthanks\repthanksunskip\def\thanks}
{}{}
\patchcmd{\@maketitle}
{\def\thanks}
{\let\repthanks\@gobble\def\thanks}
{}{}
\newcommand\repthanksunskip[1]{\unskip{}}
\makeatother

\usepackage{color}

\begin{document}
\title{EPEdit: Redefining Image Editing with Generative AI and User-Centric Design}
%
%

\author{Hoang-Phuc Nguyen\thanks{Both authors contributed equally to this research.\protect\label{X}}\inst{1,2}  \and
Dinh-Khoi Vo\repthanks{X}\inst{1,2}\orcidlink{0000-0001-8831-8846} \and
Trong-Le Do\inst{1,2} \and
Hai-Dang Nguyen\inst{1,2}\orcidlink{0000-0003-0888-8908} \and 
Tan-Cong Nguyen\inst{1,2} \and 
Vinh-Tiep Nguyen\inst{2,3}\orcidlink{0000-0003-4260-7874} \and
Tam V. Nguyen\inst{4}\orcidlink{0000-0003-0236-7992} \and
Khanh-Duy Le\inst{1,2} \and
Minh-Triet Tran\inst{1,2}\orcidlink{0000-0003-3046-3041} \and
Trung-Nghia Le\thanks{Corresponding author.}\inst{1,2}\orcidlink{0000-0002-7363-2610}}

\titlerunning{EPEdit: Redefining Image Editing}
\authorrunning{H.-P. Nguyen et al.}
%

\institute{University of Science, VNU-HCM, Ho Chi Minh City, Vietnam \and
Vietnam National University, Ho Chi Minh City, Vietnam \and
University of Information Technology, VNU-HCM, Vietnam \and
University of Dayton, Dayton, Ohio, United States\\
\email{\{nhphuc20,vdkhoi20\}@clc.fitus.edu.vn, \{dtle,nhdang,ntcong\}@selab.hcmus.edu.vn, tiepnv@uit.edu.vn, tamnguyen@udayton.edu, \{lkduy,tmtriet,ltnghia\}@fit.hcmus.edu.vn}}

\maketitle              
\begin{abstract}
The demand for image manipulation has seen a significant increase recently. Traditional tools like Photoshop and Capture One, while powerful, require considerable expertise to use effectively. Generative AI has introduced alternative platforms, such as Luminar Neo, Pixlr X, and Canva. However, many of these solutions, including resource-heavy models like Stable Diffusion, often require substantial retraining and fine-tuning, leading to high costs for users. To address these challenges, we introduce Efficient Photo Editor (EPEdit), an application that integrates a robust backend framework with a user-friendly front-end interface. EPEdit supports a wide range of creative image editing tasks, including image generation, object replacement, object removal, background modification, changes in object pose or perspective, region-specific editing, and thematic collection design, all guided by masks and prompts. Users can interact with the system through simple text commands or by marking areas for precise adjustments, making it accessible even to those without technical expertise. At its core, EPEdit leverages zero-shot image editing algorithms based on Stable Diffusion model, removing the need for additional fine-tuning. This approach enables efficient image manipulation and thematic collection creation. User evaluations for tasks of image editing, thematic design, and overall system performance demonstrate that EPEdit outperforms existing solutions, offering a user-friendly, cost-effective solution for comprehensive image editing.

\end{abstract}
\section{Introduction}

The rapid advancements in artificial intelligence (AI)~\cite{sadek2007artificial,mellit2008artificial,agah2013introduction,li2021artificial}, particularly in the realm of image manipulation, have fundamentally transformed creative practices across various domains. Central to this transformation is the emergence of Stable Diffusion (SD)~\cite{ramesh2021zeroshot,NEURIPS2021_49ad23d1,nichol2022glide,yu2022scaling,ramesh2022hierarchical,NEURIPS2022_ec795aea,rombach2022highresolution}. SD has garnered attention for its exceptional accessibility, operating efficiently on standard hardware, and its ability to generate high-quality, realistic images from textual descriptions. This versatility makes SD particularly valuable for a wide range of applications, from concept art and product design to personalized content creation. 


However, integrating SD models into current creative workflows presents challenges. Traditional softwares like Photoshop and Capture One offer robust capabilities but demand significant expertise and time investment. Modern AI-driven softwares like Luminar Neo, Pixlr X, Canva, and Midjourney have simplified image editing with user-friendly interfaces, but they often lack the deep control and flexibility professionals need. These tools require significant time and practice for users to master, and for those unfamiliar with technology or art, the learning curve can be daunting, limiting creativity. Additionally, the resource-heavy nature of AI models like SD poses challenges, particularly in environments where cost and accessibility are critical, such as educational and professional settings.


This paper addresses these issues by introducing Efficient Photo Editor 
(EPEdit), a system designed to balance advanced image manipulation with user-friendly design. EPEdit leverages SD’s strengths while reducing its resource demands via zero-short algorithms \cite{cao_2023_masactrl, 10.1145/3613905.3650788}, offering a feature-rich for generative AI-powered photo editor. We support various creative image editing features, such as image generation, object replacement, object removal, background modification, changes in object pose or perspective, region-specific editing, and thematic collection design; all guided by masks and prompts. Meanwhile, we design a simple yet efficient interface that accommodates users of all skill levels. User studies demonstrate that our solution not only enhances the creative potential of users but also addresses the practical limitations associated with current AI-driven image manipulation technologies. By making sophisticated image editing more accessible and efficient, EPEdit can empower a wider audience to explore and expand their creative possibilities.

The main contributions of this paper are as follows:

\begin{itemize}
  \item We demonstrate the system's versatility in a variety of creative image editing tasks, including region-specific edits, object replacement and removal, pose and perspective adjustments, and background alterations.
  \item We design a user-centric interface that prioritizes ease of use, making advanced functionality accessible to a broad range of users, including professionals, students, and hobbyists alike. 

  \item We conduct comprehensive user studies to evaluate the system's performance, usability, and effectiveness in real-world scenarios.
\end{itemize}

\section{Related Work}
\label{sec:AI_Tools_related}

\subsection{Modern AI-based Image Editor Tools}

Modern image editing workflows are increasingly incorporating the power of AI. These AI image editor tools leverage machine learning and generative AI to enhance and manipulate images, offering a range of benefits and some drawbacks to users. Adobe Photoshop~\footnote{\href{https://www.adobe.com/products/photoshop.html}{Adobe Photoshop}} includes Sensei AI, a suite of AI features for tasks like background removal, object selection, and intelligent upscaling, to enhance the editing process by automating repetitive tasks and improving precision. Luminar Neo~\footnote{\href{https://skylum.com/luminar-neo}{Luminar Neo}}, a standalone AI-powered photo editor, offers a range of features like noise reduction, subject enhancement, and mood adjustments based on AI algorithms. Luminar Neo simplifies complex editing tasks, making it accessible for users with varying levels of expertise. Topaz Studio~\footnote{\href{https://www.topazlabs.com/}{Topaz Studio}} provides a collection of AI-powered tools for various image editing tasks, including noise reduction, sharpening, and creative style transfers. Topaz Studio's AI capabilities help users achieve professional-grade results with minimal effort. Pixlr X~\footnote{\href{https://pixlr.com/x/}{Pixlr X}}, a free online photo editor, offers AI-powered features like background removal, object replacement, and automatic image enhancements. Pixlr X democratizes access to powerful editing tools, making them available to a broader audience without the need for expensive software. While primarily known for graphic design, Canva~\footnote{\href{https://www.canva.com/}{Canva}} also offers AI-powered features for photo editing, including background removal and photo resizing. Canva's user-friendly interface and AI capabilities make it a versatile tool for both professional designers and casual users.

\subsection{Pros and Cons of AI-based Image Editing Tools}

AI image editing tools offer several advantages. Firstly, they enhance efficiency by automating repetitive tasks like background removal, object selection, and noise reduction, freeing up users' time to focus on the creative aspects of image editing. Secondly, they improve quality as AI algorithms can analyze images and apply adjustments like color correction and sharpening with greater precision than manual techniques, leading to noticeably improved image quality. Furthermore, the accessibility of some AI image editing tools, particularly free online options, democratizes access to powerful editing capabilities that were previously exclusive to expensive professional software, thus reaching a wider range of users. Finally, AI tools like style transfer and artistic filters can inspire new creative directions and unlock possibilities for artistic expression.

However, there are some drawbacks to consider. While AI can automate tasks, users may have limited control over specific aspects of the editing process compared to traditional methods. Moreover, mastering some AI tools may require users to learn new interfaces and workflows, especially for those unfamiliar with AI-powered editing. Additionally, AI algorithms can inherit biases from the data they are trained on, leading to unintended results in image editing. Lastly, cost considerations are significant; while some free and open-source AI image editors exist, popular tools often come with subscription fees or require purchasing software licenses.



AI image editing tools offer a compelling range of benefits that can enhance efficiency, improve image quality, and unlock creative possibilities. However, it's important to be aware of limitations such as control, learning curve, and potential bias. By understanding the benefits and drawbacks, users can make informed decisions about incorporating AI tools into their image editing workflows, considering both their needs and budget.

\section{Proposed Efficient Photo Editor}

\subsection{Conceptualization}



AI has played a pivotal role in supporting human in creativity, transforming the ideas into the lifes. A particularly compelling application of AI lies in its ability to recognize, detect, and process diverse data types. AI-driven image editing tools can manipulate visual content, allowing for image generation, enhancement, and editing with minimal user input, enabling artists and designers to explore new creative possibilities. Softwares like Adobe Photoshop and AItubo have already leveraged AI models for image processing and generation; however, many of these solutions remain limited, addressing specific needs without offering a comprehensive approach for more dynamic, context-aware image editing.


\begin{figure}[t!]
    \centering
    \includegraphics[width=\textwidth]{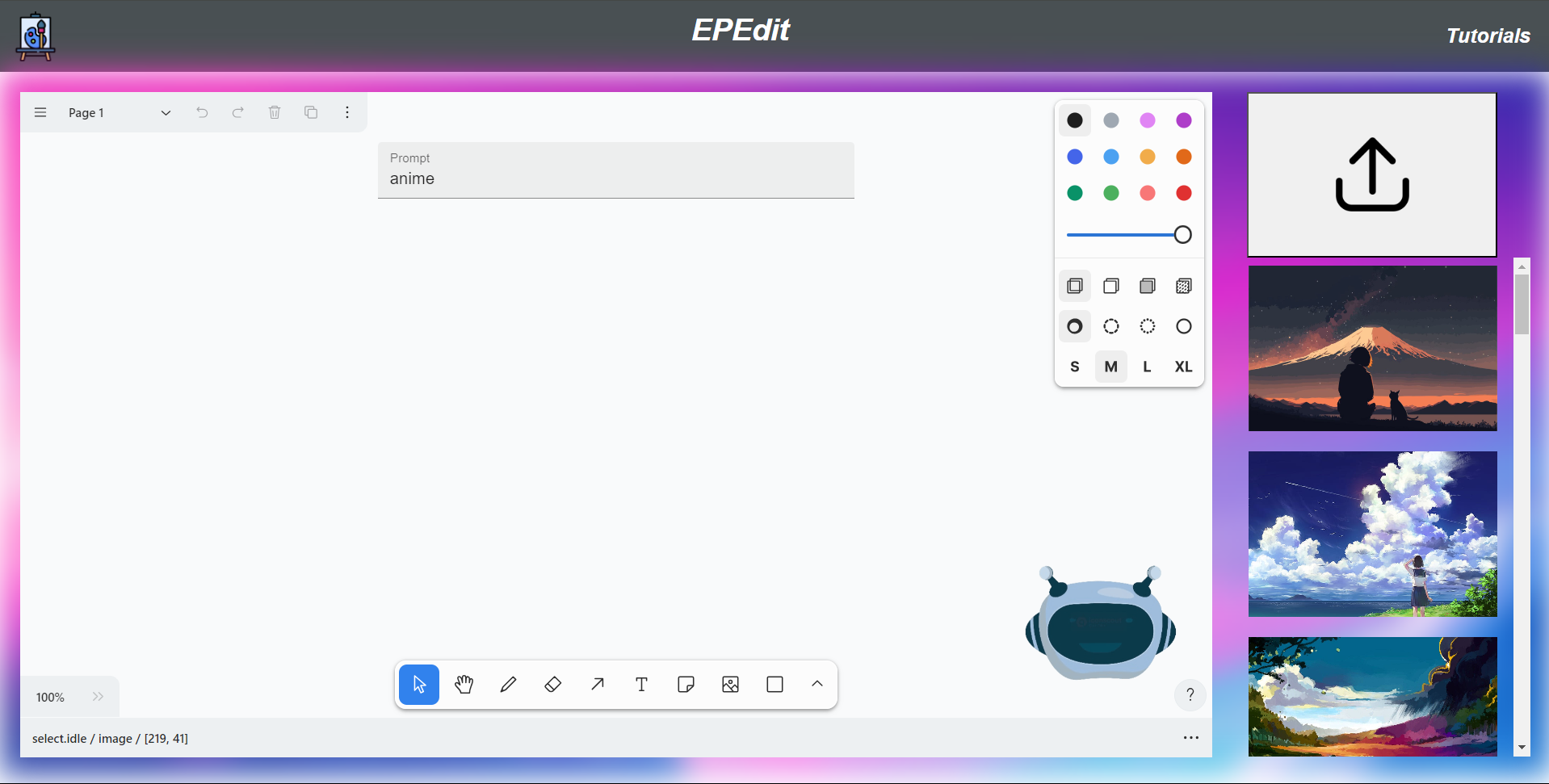}
       
      \caption{Web-based interface of the proposed EPEdit.}
      \label{fig:EPEdit}
\end{figure}

In response to the evolving needs of image processing and the limitations of existing tools, we have undertaken the development of a web-based application named Efficient Photo Editor (EPEdit), as illustrated in Fig.~\ref{fig:EPEdit}. The core objective of EPEdit is to deliver a sophisticated image-processing tool that merges efficiency with a user-friendly design. Opting for a web application format allows us to circumvent common issues associated with software installation and compatibility. Users can access EPEdit directly through their web browsers, eliminating the need for complex installations or compatibility concerns. This accessibility ensures that both seasoned developers and creative enthusiasts can engage with the system effortlessly, without encountering steep learning curves or cumbersome login processes.

\begin{figure}[t!]
    \centering
    \includegraphics[width=\textwidth]{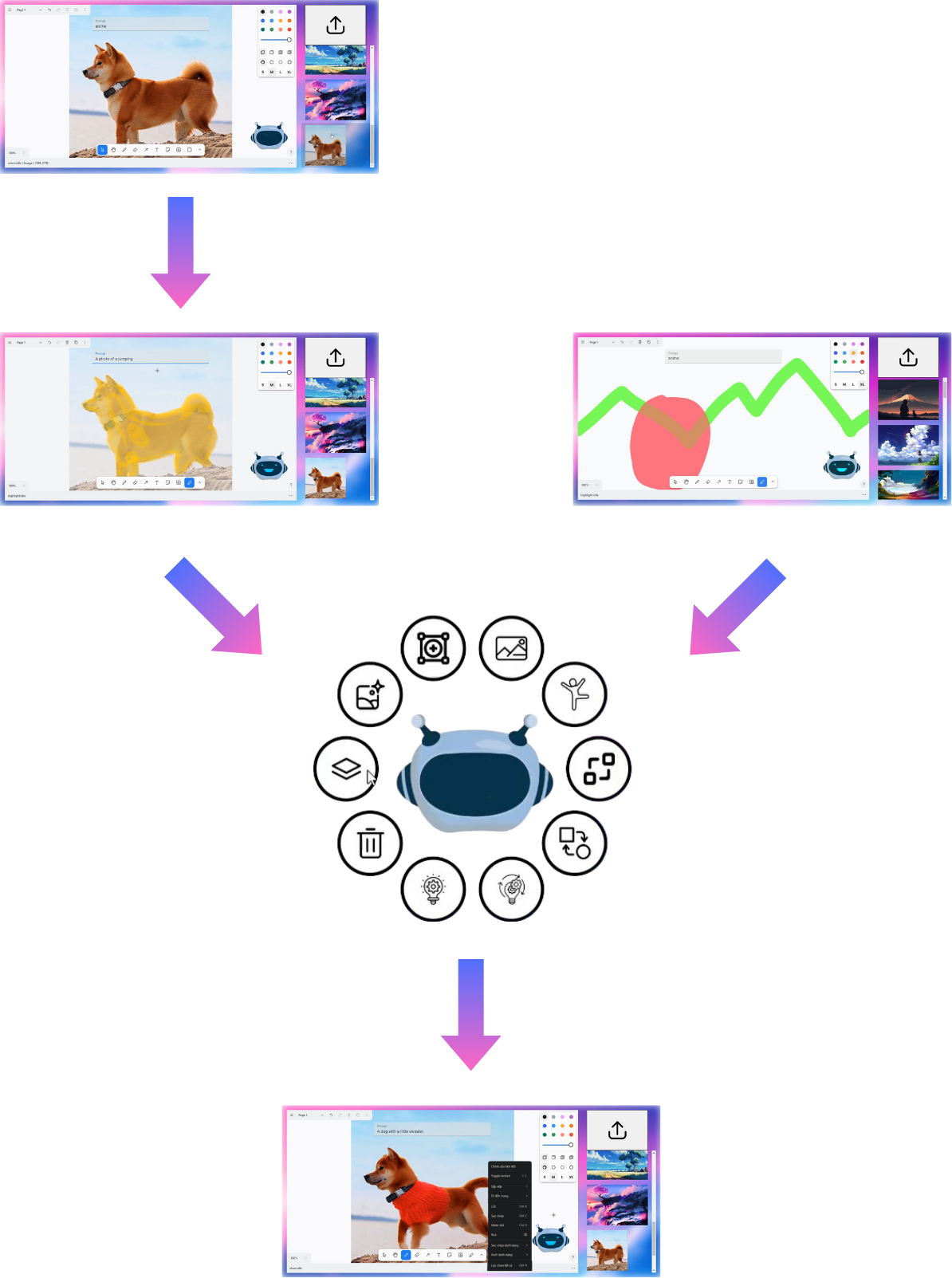}
       
      \caption{Interaction flow in EPEdit: drawing the mask, followed by image generation.}
      \label{fig:InteractionFlow}
\end{figure}

EPEdit’s underlying algorithms \cite{cao_2023_masactrl, 10.1145/3613905.3650788} are designed to perform complicated image-processing tasks with precision and speed. Whether users are enhancing photographs, creating digital art, or refining design projects, EPEdit is engineered to deliver high-quality results efficiently. Recognizing the rapid pace of technological advancement, EPEdit is developed with scalability in mind. This forward-thinking approach allows for the seamless integration of emerging AI models and the expansion of features as user needs evolve, ensuring that the application remains relevant and effective over time.

The application empowers the users to input various forms of visual content, ranging from rough sketches and concept drafts to fully developed images, into EPEdit. The result is a transformation of these inputs into vibrant, refined images that bring creative concepts to life. EPEdit is positioned to serve as a tool for designers, photographers, and hobbyists alike, promising to enhance the creative process and support the users in achieving their artistic goals.

\subsection{Graphic User Interface and Interaction Flow}

\textbf{Graphic User Interface. } We have designed an application with a simple and easy-to-use interface for the users (as detailed in Fig.~\ref{fig:EPEdit}) so that they can use it immediately upon accessing it without having to go through complicated login steps. The interface is optimized with a canvas section that is both used for drawing and processing and displaying images placed in the center of the website with a large size, making it easy for the users to manipulate. Buttons and drawing tools inside the canvas serve both drawing on the canvas and some necessary tasks for other advanced functions of the application. Next to it, a photo library contains some suggested photos. At the same time, the results of the user's photo editing functions are temporarily saved there so that the users can reuse them whenever necessary. Most notably, the virtual assistant section is located inside the canvas, the virtual assistant can suggest advanced photo editing functions to help the users easily select and perform.

\textbf{Interaction Flow. } As shown in Fig.~\ref{fig:InteractionFlow}, the users can use the application as a normal drawing application, using the drawing tools available on the canvas. In addition, the users can also upload their photos to the application or use photos available in the photo library, then use the virtual assistant feature to mask the area of the object in the photo that needs to be edited. After that, the users need to enter a description of their editing idea in the prompt input box. Finally, the users just need to choose one of the features suggested by the virtual assistant and wait for the results. After getting the results, the users can easily choose to save the results to their device if they want.

\subsection{Functionality}

\begin{figure}[t!]
    \centering
    \includegraphics[width=1.0\textwidth]{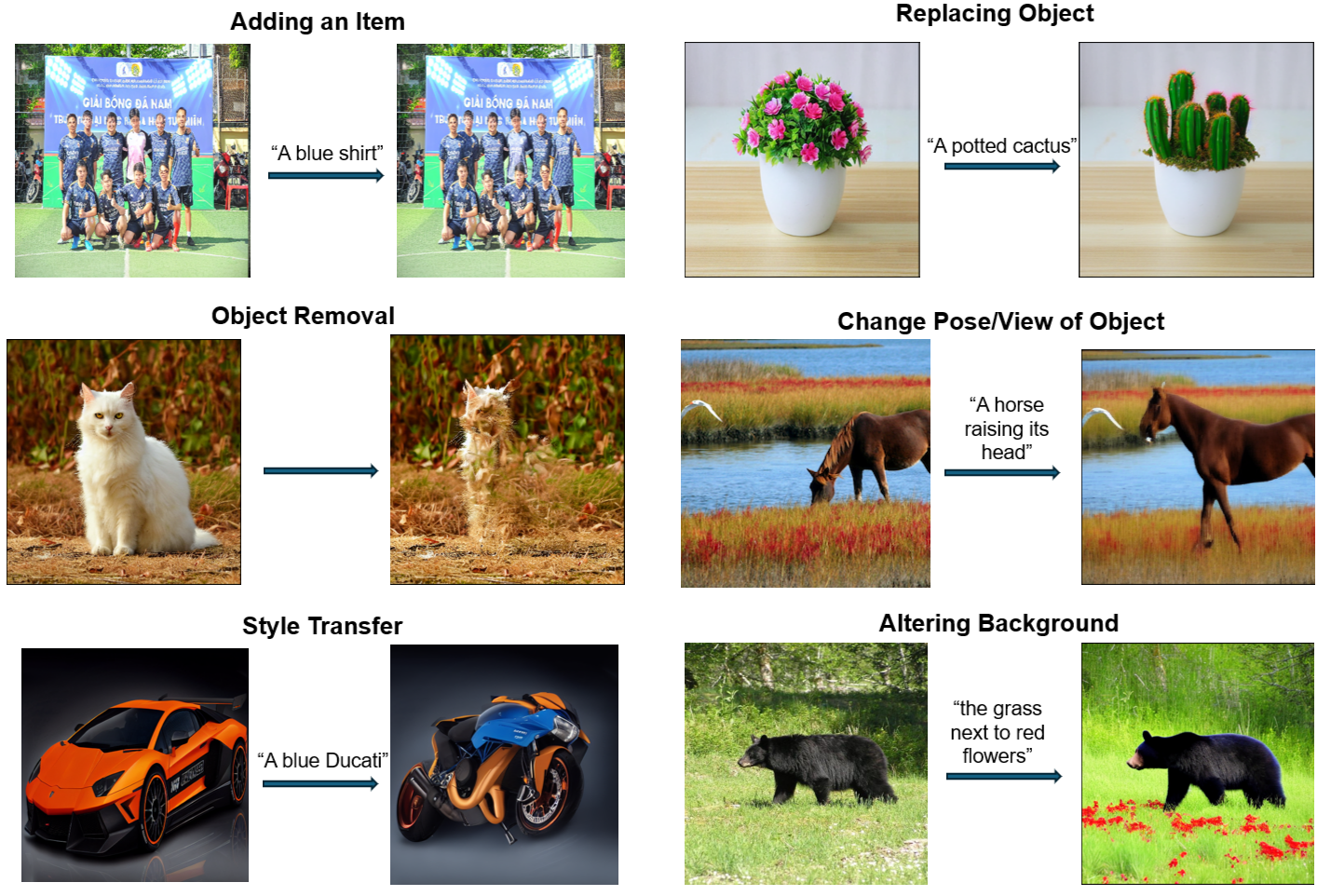}
      \caption{Multiple image editing tasks of EPEdit, such as adding an item, replacing object, style transfer, object removal, changing pose/view of object and altering the background in effortless and creative way.}
      \label{fig:Visualize_Tasks}
\end{figure}

Our EPEdit integrates outstanding functions. First of all, the drawing board with drawing support tools can help the users use EPEdit as a normal drawing tool. In addition, the users can ask the virtual assistant to help perform other advanced functions. First, the users can create images from descriptions and basic strokes drawn from available drawing tools or upload photos to the application. Then, they can mask their photos and perform editing functions such as changing an object in the photo to the object that the user describes or completely remove it from the photo easily. The users can also add a new object to the photo, or more complicatedly, change the background of that photo. Not only that, the most interesting and creative functions of EPEdit are changing the shape of the animal in the picture, or even create a set of related objects with the same theme and color as the original picture (as shown in Fig.~\ref{fig:Visualize_Tasks}). All of these are extremely creative and effective editing functions to help users freely edit images as easily as possible.

\subsection{Implementation}

\begin{figure}[t!]
    \centering
    \includegraphics[width=0.6\textwidth]{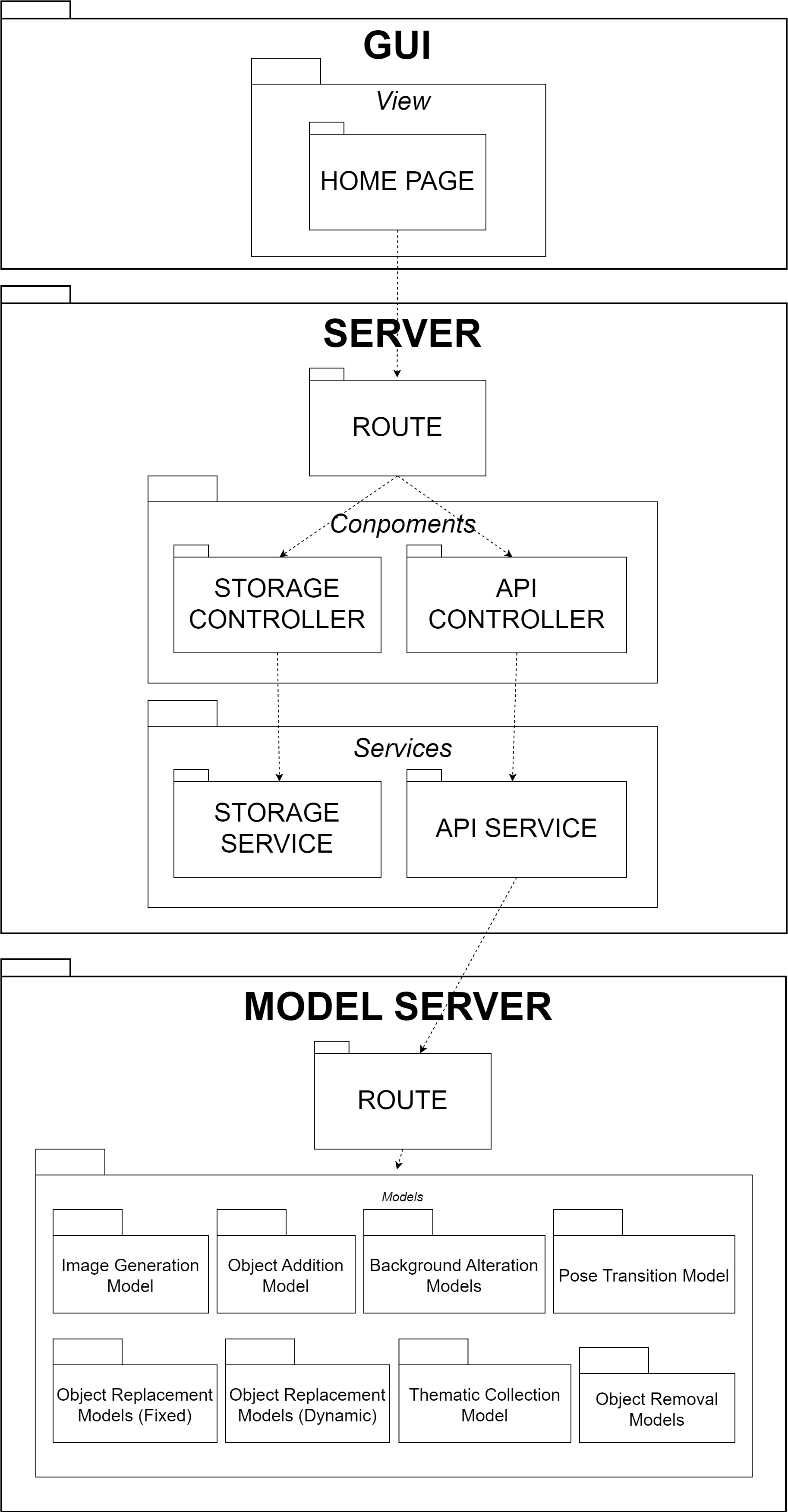}
        
      \caption{Structure of our web-based application EPEdit.}
      \label{fig:structurediagram}
\end{figure}

\textit{System Architecture. } To develop the EPEdit application, we used a popular and effective system architecture, the Model-View-Controller (MVC) architectural pattern (see Fig.~\ref{fig:structurediagram}). This architecture helps manage system components independently and allows communication between components using APIs.

\textit{Graphic User Interface. } The core of the user interface (UI) development for EPEdit is the selection of ReactJS. ReactJS is chosen for its component-based architecture, which is pivotal in enabling dynamic rendering and automatic data updates. This architectural choice ensures interactions, providing users with a highly responsive and engaging experience.

\textit{Backend. } The backend infrastructure of EPEdit is built on the Express framework. Express plays a critical role in managing the server-side operations of the application. It is responsible for handling API requests that originate from the client-side. By efficiently routing these incoming requests to the appropriate endpoints, Express ensures smooth and reliable communication between the frontend and backend components of the application.

\textit{Model Server. } To address the complex requirements of image processing, EPEdit incorporates a dedicated Python-based model server. This server is specifically designed to host pre-trained models to perform various tasks such as image generation, style transfer, and object adjustment. When the backend of the EPEdit receives an API call, the request is forwarded to the model server, typically through a post request. Upon completion of the image processing tasks by the model server, the AI-generated results are transmitted back to the backend of the EPEdit. The backend stores processed images in a gallery folder on the server, allowing users to access them anytime. It then renders these images in the GUI, providing a visual representation of the edited content.

A significant aspect of EPEdit’s architectural design is the preloading of AI models on the model server. This strategy is employed to achieve faster execution times, as the models are already equipped with the necessary libraries, thereby eliminating the need for repeated loading with each API call. This approach not only enhances the efficiency of the image processing tasks but also contributes to a smoother and more efficient user experience.

\section{User Study}
We conducted a user study to evaluate the impact and effectiveness of our proposed system, EPEdit, in practical use. Participants tried EPEdit, completed a survey, and provided an overall review. Key assessment parameters included user-friendliness, ease of use, ability to meet user requirements, and overall satisfaction. The evaluation was performed on a GPU A-100, requiring approximately 12–15 seconds per sample.

\textbf{Metrics: } We identified four key metrics to evaluate EPEdit based on user trials:
\begin{itemize}
    \item \textit{User-friendly interface}: This refers to a simple yet aesthetically pleasing design that appeals to users and features a well-organized layout, minimizing the steps required for operation. It is designed to accommodate users of varying skill levels.

    \item \textit{Ease of use}: This metric measures how easily users can navigate and become familiar with the interface. The goal is to ensure that users can quickly learn and retain the necessary actions for efficient photo editing.
    
    \item \textit{Expectation alignment}: This evaluates whether the application's features align with users' expectations, ensuring that it fully meets their photo editing needs.
    
    \item \textit{Satisfaction level}: This assesses the overall user satisfaction, focusing on whether the interface and functionalities perform well enough to meet or exceed user expectations.
\end{itemize}

\begin{figure}[!t]
    \centering
    \includegraphics[width=\textwidth]{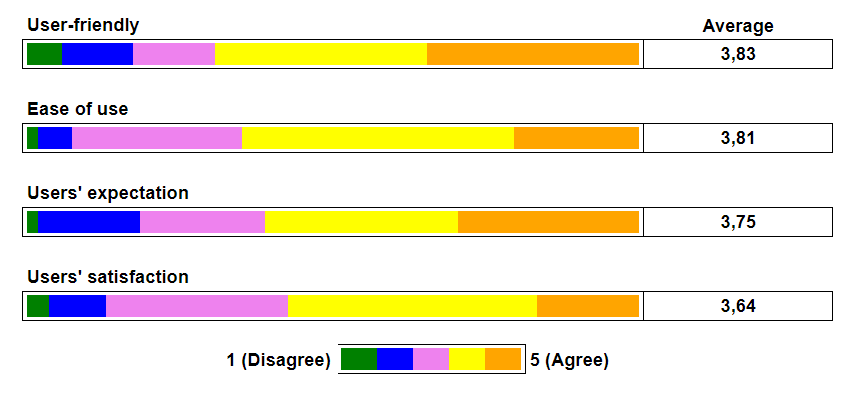}
      
      \caption{Quantitative data of the user study on EPEdit for creative image editing tasks.}
     \label{fig:userstudy-chart01}
\end{figure}

\textbf{Participants: } We invited 24 volunteers to participate in our open survey (15 men, and 9 women; all are from 20 to 30). The participants including students studying software engineering and AI and others with diverse expertise.

\textbf{Apparatus and procedure: } 
The user study was conducted on-site in our laboratory. Users experienced our web-based system in some design tasks and then filled out the evaluation form. Our survey administration was conducted online, which took about 15 minutes per person.

Participants were asked to try out our proposed system on certain test functions and then rated their level of agreement on a scale from 1 (Disagree) to 5 (Agree) indicators based on their views. They experienced features such as import/export, image generation, object removal, object addition, and background alteration.

They also tried to used Photoshop in order to compare with our proposed system. For fair comparison, those who have not used Photoshop watched a short tutorial about how to use Photoshop so they can take a quick look and make an immediate contents.

\textbf{Quantitative results: } The user study results in Fig.~\ref{fig:userstudy-chart01} shows that our proposed system met the users' expectations and satisfaction in supporting creative image editing tasks. The participants also highly evaluated the user-friendly interface and ease of use of EPEdit.

\begin{figure}[!t]
    \centering
    \begin{subfigure}[t]{0.45\textwidth}
        \centering
        \includegraphics[width=\textwidth]{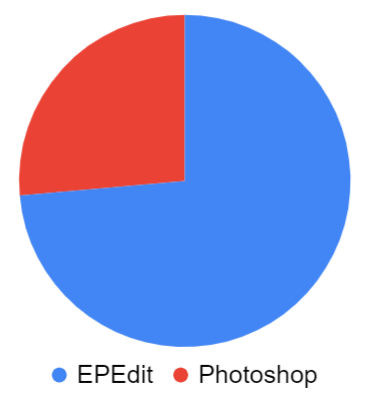}
        \caption{Ease of use}
        \label{fig:convenience}
    \end{subfigure}
    ~ 
    \begin{subfigure}[t]{0.45\textwidth}
        \centering
        \includegraphics[width=\textwidth]{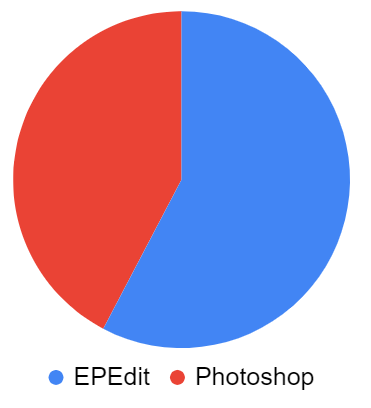}
        \caption{Speed and performance}
        \label{fig:speed&performance}
    \end{subfigure}
        \caption{Comparative performance of EPEdit and Photoshop in creative image editing. Our proposed system outperforms Photoshop.}
        \label{fig:userstudy-chart02}
\end{figure}

When compared with one of the current leading photo editing softwares - Photoshop, our EPEdit significantly outperforms Photoshop in both easy of use as well as speed and quality of results (see Fig.~\ref{fig:userstudy-chart02}). Unlike Photoshop requiring high expertise, our system can be used by anyone and directly supports creative image editing tasks.

\section{Conclusion and Future Work}
In this paper, we have introduced EPEdit, a photo editing application on the web platform, with a friendly interface and easy to access, easy to use. Our proposed system utilizes AI models for supporting users to creatively edit photos quickly. We have conducted the user study to verify the effectiveness, feasibility, and accessibility of our EPEdit in support anyone to edit photos easily.

Based on the feedback from participants in the user study, we plan to integrate new algorithms and new features in the future to diversify creativity and meet the needs of users in photo editing. In addition, we find that upgrading the ability of the masking function for objects in the application's photos is also necessary. In addition, assisting users to enter prompt descriptions will be more convenient via voice instead of typing manually.

\section*{Acknowledgements} 

This research is funded by Vietnam National Foundation for Science and Technology Development (NAFOSTED) under Grant Number 102.05-2023.31. 

%
%

\bibliographystyle{splncs04}
\bibliography{mybibliography}

@String{Computing = "Computing" }

@String{Computer = "{IEEE} Computer" }

@String{Chelsea = "Chelsea" }

@String(ICCV= {Int. Conf. Comput. Vis.})

@String(ICCV  = {ICCV})

@inproceedings{cao_2023_masactrl,
    author    = {Cao, Mingdeng and Wang, Xintao and Qi, Zhongang and Shan, Ying and Qie, Xiaohu and Zheng, Yinqiang},
    title     = {MasaCtrl: Tuning-Free Mutual Self-Attention Control for Consistent Image Synthesis and Editing},
    booktitle = {Proceedings of the IEEE/CVF International Conference on Computer Vision (ICCV)},
    month     = {October},
    year      = {2023},
    pages     = {22560-22570}
}

@article{nichol2022glide,
  title={Glide: Towards photorealistic image generation and editing with text-guided diffusion models},
  author={Nichol, Alex and Dhariwal, Prafulla and Ramesh, Aditya and Shyam, Pranav and Mishkin, Pamela and McGrew, Bob and Sutskever, Ilya and Chen, Mark},
  journal={arXiv preprint arXiv:2112.10741},
  year={2021}
}

@article{yu2022scaling,
  title={Scaling autoregressive models for content-rich text-to-image generation},
  author={Yu, Jiahui and Xu, Yuanzhong and Koh, Jing Yu and Luong, Thang and Baid, Gunjan and Wang, Zirui and Vasudevan, Vijay and Ku, Alexander and Yang, Yinfei and Ayan, Burcu Karagol and others},
  journal={arXiv preprint arXiv:2206.10789},
  volume={2},
  number={3},
  pages={5},
  year={2022}
}

@article{ramesh2022hierarchical,
  title={Hierarchical text-conditional image generation with clip latents. arXiv 2022},
  author={Ramesh, Aditya and Dhariwal, Prafulla and Nichol, Alex and Chu, Casey and Chen, Mark},
  journal={arXiv preprint arXiv:2204.06125},
  year={2022}
}

@inproceedings{10.1145/3613905.3650788,
author = {Vo, Dinh-Khoi and Ly, Duy-Nam and Le, Khanh-Duy and Nguyen, Tam V. and Tran, Minh-Triet and Le, Trung-Nghia},
title = {iCONTRA: Toward Thematic Collection Design Via Interactive Concept Transfer},
year = {2024},
isbn = {9798400703317},
publisher = {Association for Computing Machinery},
address = {New York, NY, USA},
url = {https://doi.org/10.1145/3613905.3650788},
doi = {10.1145/3613905.3650788},
booktitle = {Extended Abstracts of the CHI Conference on Human Factors in Computing Systems},
articleno = {382},
numpages = {8},
location = {, Honolulu, HI, USA, },
series = {CHI EA '24}
}

@article{sadek2007artificial,
  title={Artificial intelligence applications in transportation},
  author={Sadek, Adel W},
  journal={Transportation research circular},
  pages={1--7},
  year={2007},
  publisher={Citeseer}
}

@article{mellit2008artificial,
  title={Artificial intelligence techniques for photovoltaic applications: A review},
  author={Mellit, Adel and Kalogirou, Soteris A},
  journal={Progress in energy and combustion science},
  volume={34},
  number={5},
  pages={574--632},
  year={2008},
  publisher={Elsevier}
}

@article{agah2013introduction,
  title={Introduction to medical applications of artificial intelligence},
  author={Agah, Arvin},
  journal={Medical Applications of Artificial Intelligence},
  pages={19--26},
  year={2013},
  publisher={CRC Press}
}

@article{li2021artificial,
  title={Artificial intelligence applications in psychoradiology},
  author={Li, Fei and Sun, Huaiqiang and Biswal, Bharat B and Sweeney, John A and Gong, Qiyong},
  journal={Psychoradiology},
  volume={1},
  number={2},
  pages={94--107},
  year={2021},
  publisher={Oxford University Press}
}

@inproceedings{ramesh2021zeroshot,
  title={Zero-shot text-to-image generation},
  author={Ramesh, Aditya and Pavlov, Mikhail and Goh, Gabriel and Gray, Scott and Voss, Chelsea and Radford, Alec and Chen, Mark and Sutskever, Ilya},
  booktitle={International conference on machine learning},
  pages={8821--8831},
  year={2021},
  organization={Pmlr}
}

@inproceedings{rombach2022highresolution,
  title={High-resolution image synthesis with latent diffusion models},
  author={Rombach, Robin and Blattmann, Andreas and Lorenz, Dominik and Esser, Patrick and Ommer, Bj{\"o}rn},
  booktitle={Proceedings of the IEEE/CVF conference on computer vision and pattern recognition},
  pages={10684--10695},
  year={2022}
}

@inproceedings{NEURIPS2022_ec795aea,
 author = {Saharia, Chitwan and Chan, William and Saxena, Saurabh and Li, Lala and Whang, Jay and Denton, Emily L and Ghasemipour, Kamyar and Gontijo Lopes, Raphael and Karagol Ayan, Burcu and Salimans, Tim and Ho, Jonathan and Fleet, David J and Norouzi, Mohammad},
 booktitle = {Advances in Neural Information Processing Systems},
 editor = {S. Koyejo and S. Mohamed and A. Agarwal and D. Belgrave and K. Cho and A. Oh},
 pages = {36479--36494},
 publisher = {Curran Associates, Inc.},
 title = {Photorealistic Text-to-Image Diffusion Models with Deep Language Understanding},
 url = {https://proceedings.neurips.cc/paper_files/paper/2022/file/ec795aeadae0b7d230fa35cbaf04c041-Paper-Conference.pdf},
 volume = {35},
 year = {2022}
}

@inproceedings{NEURIPS2021_49ad23d1,
 author = {Dhariwal, Prafulla and Nichol, Alexander},
 booktitle = {Advances in Neural Information Processing Systems},
 editor = {M. Ranzato and A. Beygelzimer and Y. Dauphin and P.S. Liang and J. Wortman Vaughan},
 pages = {8780--8794},
 publisher = {Curran Associates, Inc.},
 title = {Diffusion Models Beat GANs on Image Synthesis},
 url = {https://proceedings.neurips.cc/paper_files/paper/2021/file/49ad23d1ec9fa4bd8d77d02681df5cfa-Paper.pdf},
 volume = {34},
 year = {2021}
}

\end{document}